\pdfoutput=1

\documentclass[11pt]{article}

\usepackage[]{acl}

\usepackage{times}
\usepackage{latexsym}

\usepackage[T1]{fontenc}

\usepackage[utf8]{inputenc}

\usepackage{microtype}

\usepackage{amsmath}
\usepackage{amssymb}
\usepackage{graphicx,paralist,multirow}
\usepackage{booktabs}
\usepackage[inline]{enumitem}

\title{Structured Dialogue Discourse Parsing}

\author{Ta-Chung Chi \\
    Language Technologies Institute \\
  Carnegie Mellon University \\
  \texttt{tachungc@andrew.cmu.edu} \\\And
  Alexander I. Rudnicky \\
  Language Technologies Institute \\
  Carnegie Mellon University \\
  \texttt{air@cs.cmu.edu} \\}

\begin{document}
\maketitle
\begin{abstract}
Dialogue discourse parsing aims to uncover the internal structure of a multi-participant conversation by finding all the discourse~\emph{links} and corresponding~\emph{relations}.
Previous work either treats this task as a series of independent multiple-choice problems, in which the link existence and relations are decoded separately, or the encoding is restricted to only local interaction, ignoring the holistic structural information. In contrast, we propose a principled method that improves upon previous work from two perspectives: encoding and decoding. From the encoding side, we perform structured encoding on the adjacency matrix followed by the matrix-tree learning algorithm, where all discourse links and relations in the dialogue are jointly optimized based on latent tree-level distribution. From the decoding side, we perform structured inference using the modified Chiu-Liu-Edmonds algorithm, which explicitly generates the labeled multi-root non-projective spanning tree that best captures the discourse structure. In addition, unlike in previous work, we do not rely on hand-crafted features; this improves the model's robustness. Experiments show that our method achieves new state-of-the-art, surpassing the previous model by 2.3 on STAC and 1.5 on Molweni (F1 scores). \footnote{Code released at~\url{https://github.com/chijames/structured_dialogue_discourse_parsing}.}
\end{abstract}

\section{Introduction}
\begin{figure}[]
\hspace{1em}\includegraphics[width=0.4\textwidth]{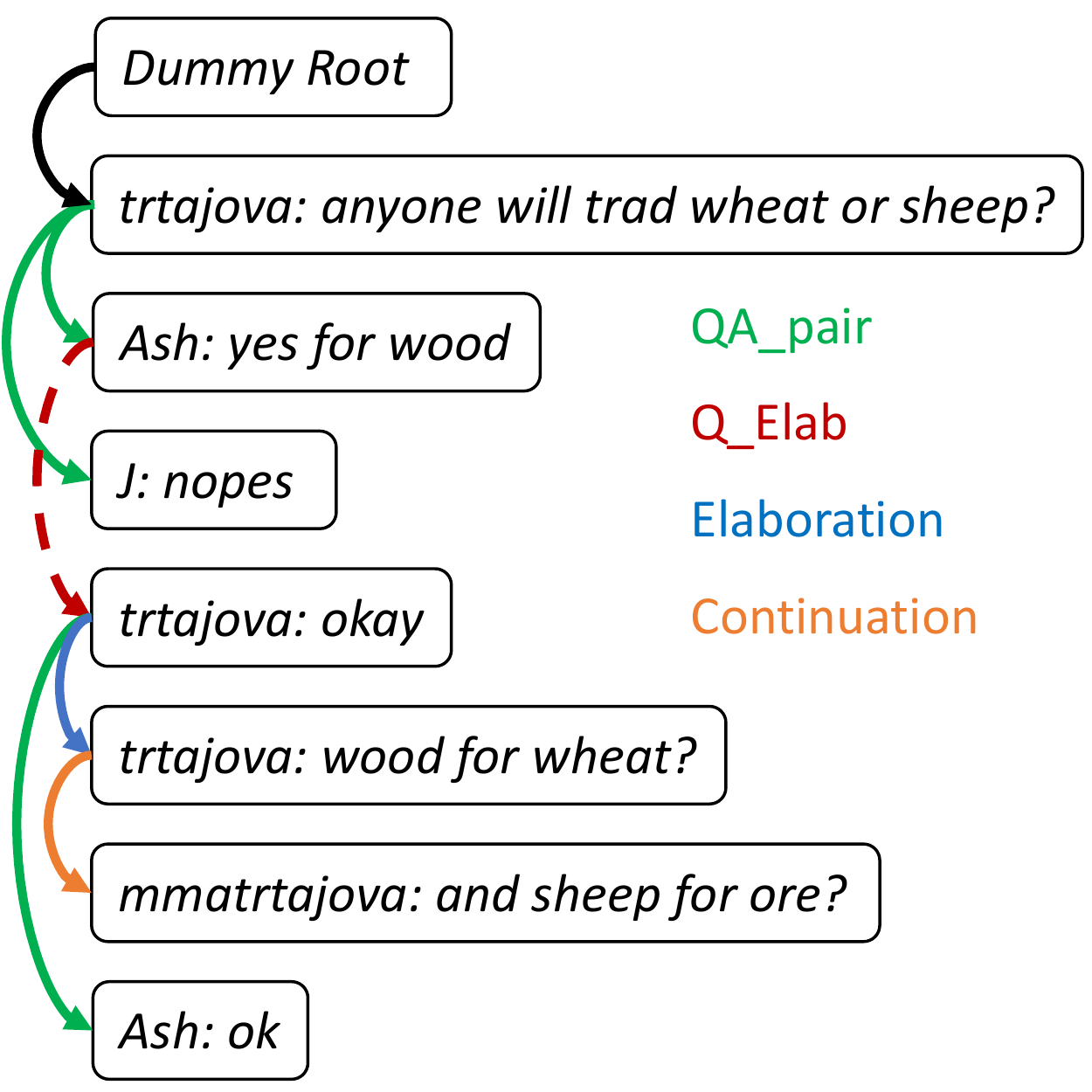}
\caption{This is an example dialogue session. The ultimate goal of a dialogue discourse parser is to predict all the links (arrows) and relations (color of arrows) shown in this figure. Note that the Q\_Elab arrow (dashed) can cross the QA\_pair one, making it non-projective.}
\label{fig:task}
\vspace{-3mm}
\end{figure}

Discourse parsing is a series of tasks that consist of elementary discourse unit (EDU) segmentation, relation directionality classification (optional), and relation type classification between EDUs~\cite{jurafsky_martin_2021}. It serves as the first step of many downstream applications~\cite{meyer2012using, jansen2014discourse,narasimhan2015machine,bhatia2015better,ji2016latent,asher2016discourse, ji2017neural, li2020molweni},
and it can be categorized into three major discourse formalisms: RST~\cite{mann1988rhetorical}, PDTB~\cite{prasad2008penn}, and SDRT~\cite{lascarides2008segmented} styles.
Considering that SDRT-style formalism is used to label the STAC~\cite{asher2016discourse} and Molweni~\cite{li2020molweni} dialogue corpora and the increasing importance of dialogue discourse parsers trained on them~\cite{ouyang2020dialogue,feng2020dialogue,jia2020multi,chen2021structure}, we focus on designing an SDRT-style dialogue discourse parser using the two corpora in this work.
Figure~\ref{fig:task} presents an example of a dialogue session in the STAC corpus~\cite{asher2016discourse} annotated with its discourse structure.
The annotation is often encoded in two components:~\emph{links} and \emph{relations}. The goal of a dialogue discourse parser is to extract them accurately at the same time.

\begin{table*}[ht!]
\resizebox{\textwidth}{!}{%
\begin{tabular}{@{}lllcc@{}}
\toprule
\textbf{Models} & \textbf{Encoding} & \textbf{Decoding} & \multicolumn{1}{l}{\textbf{Link \& Relation Prediction}} & \multicolumn{1}{l}{\textbf{Use Feature}} \\ \midrule
MST~\shortcite{afantenos2015discourse} & \textit{local, edge-wise} & \textit{partial MST} & \textit{separate} & Y \\
ILP~\shortcite{perret2016integer} & \textit{local, edge-wise} & \textit{ILP} & \textit{separate} & Y \\
Deep-Seq.~\shortcite{shi2019deep} & \textit{global, two-staged} & \textit{indp. multiple choice} & \textit{separate} & Y \\

Struct-Aware~\shortcite{wangstructure} & \textit{global, fully-connected} & \textit{indp. multiple choice} & \textit{separate} & Y \\
Hierarchical~\shortcite{liu-chen-2021-improving} & \textit{hierarchical} & \textit{indp. multiple choice} & \textit{separate} & N \\
This Work & \textit{global, structured} & \textit{full MST} & \textit{joint} & N \\ \bottomrule
\end{tabular}%
}
\caption{This is the comparison between different dialogue discourse parsers. Our method is designed with structured encoding and decoding processes. Furthermore, the links and relations are learned and predicted jointly. Finally, our method does not rely on human-designed features, hence enjoys better robustness.}
\label{tab:comparison}
\end{table*}

One straightforward solution to this problem is to transform the parsing structure into a series of local pairwise link prediction problems. In other words, the model is expected to compute some local potentials between each pair of utterances, and predict the relation type of that link if it exists. However, this formulation does not take the global structural information into account, leading to inferior parsing performance.

In contrast to previous work, our core observation is that by adding a dummy utterance at the beginning of the dialogue, the overall structure closely resembles a \textbf{labeled multi-root non-projective spanning tree}.
In light of this observation, we propose a principled dialogue discoursing parser that encodes structural inductive biases during training and inference.

The essential elements of our method are the novel structured parameterization of the adjacency matrix, the directed version of matrix-tree theorem~\cite{tutte1984graph,koo2007structured}, and the modified directed spanning tree inference algorithm. To the best of our knowledge, this is the first time that the labeled multi-root non-projective spanning tree is applied to the analysis of dialogue discourse structure.
In summary, the contributions of this paper are:
\begin{compactitem}
    \item We propose a principled method for the dialogue discourse parsing task, where structural inductive biases for both encoding and decoding processes are introduced.
    \item We jointly predict discourse \emph{links} and \emph{relations} in a unified space.
    \item We propose a padding method that allows batchwise variable-length determinant calculation.
    \item Experimental results demonstrate state-of-the-art discoursing parsing performance on two datasets.
\end{compactitem}

\section{Task Background}
\label{sec:task}
We are given a dialogue session $D$ and the links and relations between pairs of utterances labeled using the 17 discourse relations defined in~\citet{asher2016discourse}. All the utterances, links, and relations constitute a graph $G(V, E, R)$, where $V$ represents the set of utterances, $E$ represents the links connecting them, and $R$ represents the edge labels. The goal of a discourse parser is to predict $E$ and $R$ given $V$.

There are five existing dialogue discourse parsers to the best of our knowledge~\cite{afantenos2015discourse,perret2016integer,shi2019deep,wangstructure,liu-chen-2021-improving}. We compare them against each other in detail in the following subsections and provide a summary in Table~\ref{tab:comparison}.

\subsection{Encoding}
\citet{afantenos2015discourse,perret2016integer} use a MaxEnt~\cite{ratnaparkhi1997simple} model to parameterize local pairwise scores between utterance pairs. Therefore, global and contextual information are not taken into account during the encoding process.~\citet{liu-chen-2021-improving} improve upon them by using a hierarchical encoder that models the contextual information.
\citet{shi2019deep} inject more structural information by first predicting all the links, followed by a global structured encoding module. However, the predicted links are discrete, making this two-staged solution not end-to-end trainable. To connect the two stages, \citet{wangstructure} instead use a fully connected graph between all utterances. While being fully end-to-end, useful structured bias is not encoded anymore. Based on the drawbacks of previous parsers, we propose a fully end-to-end encoder while maintaining structured information at the same time.

\subsection{Decoding}
\citet{shi2019deep,wangstructure,liu-chen-2021-improving} treat the links and relations decoding tasks as a series of independent multiple-choice problems. In other words, the existence of one link has nothing to do with other links. In contrast,~\citet{perret2016integer} find the structure by solving an integer linear programming problem, but it needs a set of complicated human-designed decoding constraints.
~\citet{afantenos2015discourse} is the closest approach to this work, where they run the maximum spanning tree decoding algorithm on the predicted edges only to find the tree structure (links). However, the relations are not jointly decoded. Instead, we run the modified spanning tree decoding algorithm on the unified link and relation space.

\subsection{Link and Relation Prediction}
All previous work treat the prediction of links and relations as a two-stage process. That is, they first predict the existence of a link, and the relation is predicted only if the link exists. This decouples the joint learning of links and relations. We mitigate this issue by unifying the prediction space of links and relations, making it a three-dimensional tensor.

\subsection{Feature Usage}
\label{sec:feature}
Finally, all previous work execpt~\cite{liu-chen-2021-improving} utilize some hand-crafted features. To name a few, they explicitly model if two utterances are spoken by the same~\emph{speaker}, or if they belong to the same~\emph{turn}. These features are useful but also make the baseline parsers deeply coupled with them, which might limit the parsing performance if applied to a new dataset. For example, if the new dataset is a transcript of a teleconference or radio exchange, it is likely that we only have the utterances recorded as it is expensive and hard to obtain all the speaker and turn information. In contrast, since our model does not rely on such explicitly modeled feature, the performance drop is less than the ones that use them when the speaker and turn information are removed.

\section{Structure Formulation}
\label{sec:structure}
The graph $G$ defined in \S~\ref{sec:task} can theoretically be any directed acyclic graph, which is generally difficult to optimize. Fortunately, we find that by discarding only a small fraction of the edges, which is 6\% for the STAC corpus and 0\% for the molweni corpus, we can recover a spanning tree-like structure that permits efficient learning and structure inference.
For the nodes having more than one parent, we keep only the latest one.\footnote{This strategy is adopted by all baselines as well.} In addition, for dangling utterances that do not have any parents, we connect them to the dummy root utterance, so we are in fact optimizing a \emph{multi-root} tree during training time. Finally, note that our tree structure allows different links to \emph{cross} each other (Figure~\ref{fig:task}) and each edge also has a relation \emph{label}, $G(V, E, R)$ is a labeled directed multi-root non-projective spanning tree, which is referred to as tree for conciseness hereinafter~\footnote{There are four types of spanning trees investigated in the dependency parsing domain~\cite{mcdonald2005non,koo2007structured}}.

Several questions naturally arise:
\begin{compactitem}
    \item How to parametrize the tree? We will model the pairwise potential scores by an adjacency matrix, where a cell represents the relevance score of a pair of utterances. See \S~\ref{sec:parameterization}.
    \item How to learn the correct tree? We calculate the probability of the correct tree among all possible trees encoded by the adjacency matrix, and that probability is maximized. This is similar to softmax attention using trees as basic units instead of tokens. See \S~\ref{sec:learning}.
    \item How to perform inference? Given the learned three-dimensional adjacency matrix, we can run the modified maximum spanning tree induction algorithm to induce the tree structure. See \S~\ref{sec:inference}.
\end{compactitem}

\begin{figure*}[]
\vspace{-1cm}
\includegraphics[width=\textwidth]{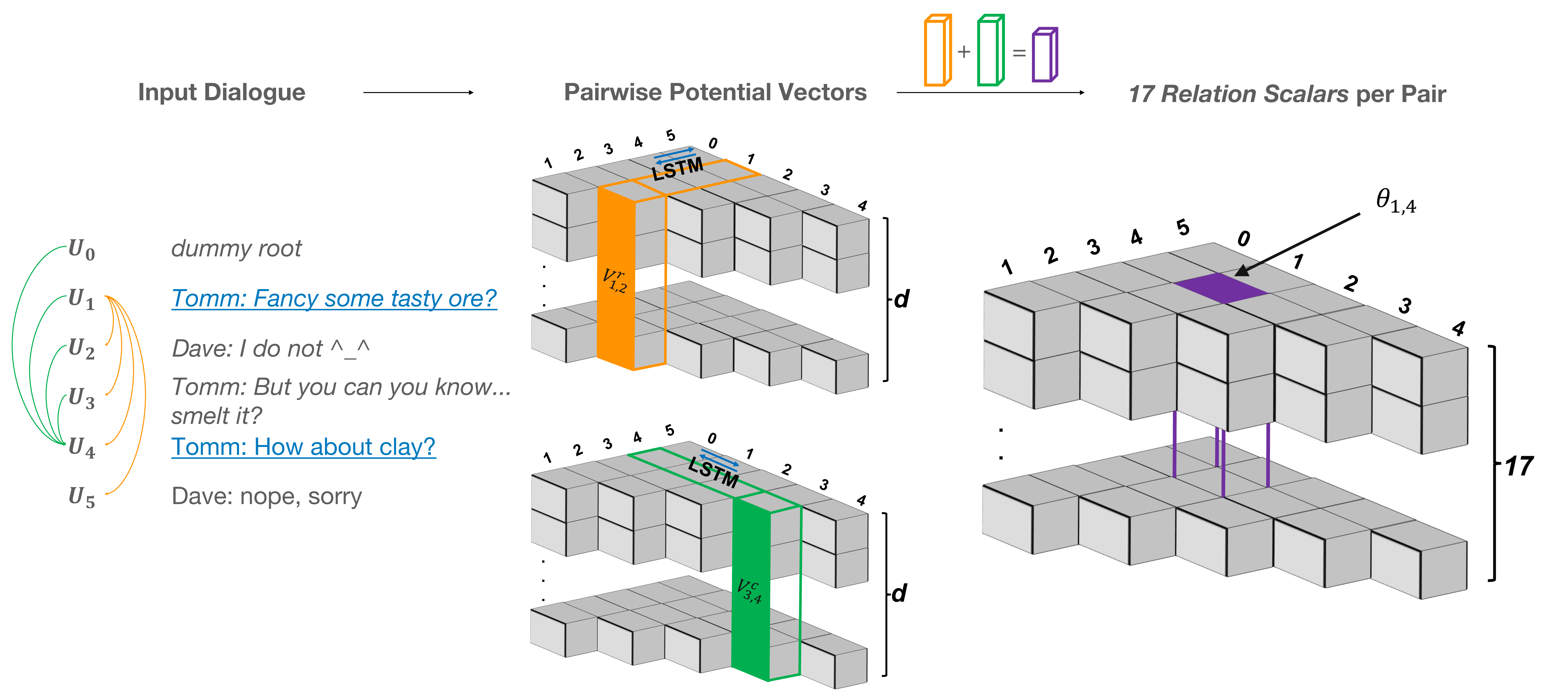}
\caption{The contextual encoding process. Row numbers from 0 to 4 represent $h$, and column numbers from 1 to 5 represents $m$. In this example, the 1-st utterance can connect to the 2-nd, 3-rd or 5-th utterances when predicting the $V_{1,4}$ cell. This is represented by the orange rectangles. Similarly, the 4-th utterance can have 0, 1, or 3-rd utterances as its parent, represented by the green rectangles. We use two LSTMs for two directions (orange and green). Finally, we add the contextualized vectors together to get the purple vector $\textrm{Linear}(V^r_{1,4}+V^c_{1,4})=\theta_{1,4}$.}
\label{fig:contextual_encoding}
\end{figure*}

\section{Proposed Framework}
\label{sec:framework}
\subsection{Model Parameterization}
\label{sec:parameterization}
Given $n$ utterances (aka EDUs) $\{U_i\}_{i=1}^n$ in a dialogue session $D$, we define a~\emph{discourse pair} in $D$ as a 3-tuple $(h,m,r),h < m, r\in [1,17]$ where $h\in [0\dots n]$ is the index of the parent utterance, $m\in [1 \dots n]$ is the index of the children utterance, and $r$ is one of the 17 relations. Note that we add a special~\emph{root} utterance $h=0$ to be the shared pseudo parent for the first utterances. Note that this root utterance can be chosen arbitrarily, and we use the utterance ``This is the start of a dialogue'' in this work.

To parameterize the tree, we first model the $d$-dimensional pairwise representation between a pair of utterances. This can be expressed compactly by a 3-order tensor,
which is an adjacency matrix with each element $V_{h,m}$ being a d-dimensional feature vector, hence
$V\in \mathbb{R}^{(n+1)\times (n+1)\times d}$.
Each $V_{h,m}$ is calculated using a BERT~\cite{devlin2018bert} model as the encoder. 
BERT takes a pair of utterances as input, and a special [CLS] token is prepended before the concatenation of the two utterances. The representation of the [CLS] token is further used to calculate $d$-dimensional pairwise representation:
\begin{equation}
    V_{h,m} = \mathrm{BERT}_{CLS}(U_h,U_m)
    \label{eq:independent_score}
\end{equation}
One immediate drawback of eq.~(\ref{eq:independent_score}) is that the pairwise scores are calculated independently. We alleviate this issue by using a bidirectional LSTM to encode the contextual information.
For the $h$-th row, we obtain the hidden states for all timestep $t$:
\begin{equation}
    \{V_{h,t}^\mathrm{r}\}_{t=h+1}^n = \mathrm{LSTM}(\{V_{h,t}\}_{t=h+1}^n)
\end{equation}
The underlying idea is that to accurately decide if $U_h$ should point to $U_m$, we should collect the information of connecting $U_h$ to all the other utterances that appear later chronologically.
Similarly, for the $m$-th column, all the hidden states are:
\begin{equation}
    \{V_{t,m}^\mathrm{c}\}_{t=0}^{m-1} = \mathrm{LSTM}(\{V_{t,m}\}_{t=0}^{m-1})
\end{equation}
The final context-aware potential score is:
\begin{equation}
    \tilde V_{h,m} = V^\mathrm{r}_{h,m}+V^\mathrm{c}_{h,m}
    \label{eq:encoder}
\end{equation}
$\tilde V\in \mathbb{R}^{(n+1)\times (n+1)\times 2d}$.
Every pairwise score is now aware of neighboring pairs.
It still remains to convert $\tilde V$ to individual score of a discourse pair. We do so by simply passing $\tilde V$ through a linear transformation layer:
\begin{equation}
    \theta_{h,m} = \mathrm{Linear}(\tilde V)
    \label{eq:encoder}
\end{equation}where
$\theta\in \mathbb{R}^{(n+1)\times (n+1)\times 17}$.
Note that there is no activation function after the linear layer since we assume $\theta$ to be in log space. Another important property of $\theta$ is that it is a strictly \emph{upper-triangular} matrix due to the $h<m$ constraint. In practice, this can be enforced by setting the lower triangular and diagonal elements to -inf. We illustrate the overall idea in Figure~\ref{fig:contextual_encoding}.

\begin{figure*}[]
\vspace{-0.8cm}
\includegraphics[width=1\textwidth]{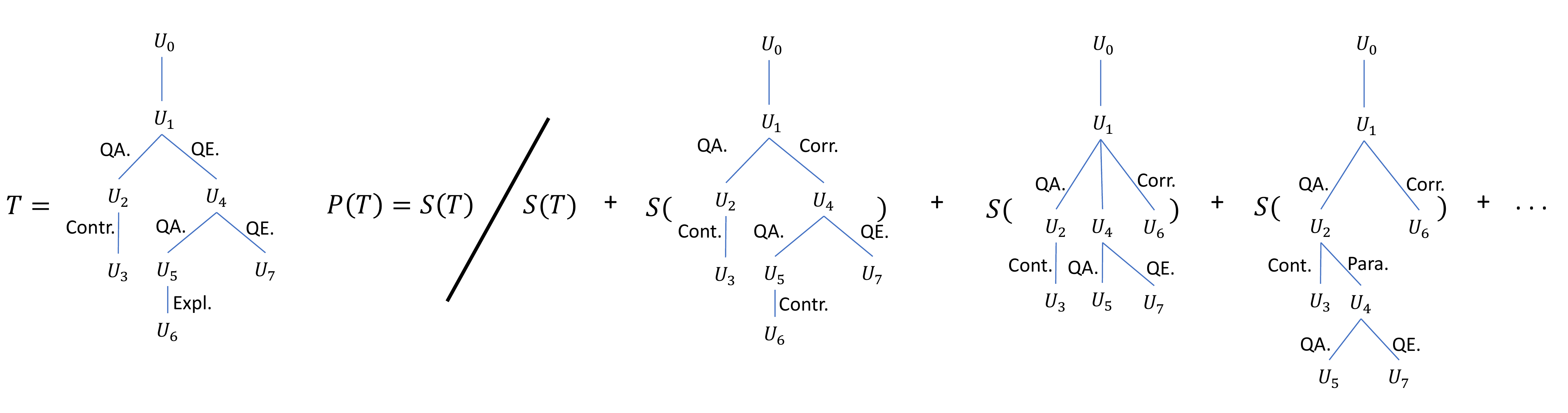}
\vspace{-0.8cm}
\caption{This is the graphical illustration of of how we perform tree-structured learning. Note that we are summing the scores of all possible labeled trees as the denominator, which is eq.~(\ref{eq:det}).}
\label{fig:tree_encoding}
\end{figure*}

\subsection{Learning the Tree}
\label{sec:learning}
The parameterization we define in eq.~(\ref{eq:encoder}) still does not impose any structural constraints. Based on our conclusion in \S~\ref{sec:structure}, we would like to impose a non-projective multi-root spanning tree constraint in the learning and inference process.
We define a tree $T$ to be the collection of discourse pairs $\{(h,m,r)\}$.
We use $\mathcal{T}(D)$ to denote all possible trees of a dialogue session $D$. During learning, the reference tree structure $\bar T\in \mathcal{T}(D)$ is given. If the score of $\bar T$ and the summation of the scores of all trees in $\mathcal{T}(D)$ is tractable, we can obtain the probability of $\bar T$ and optimize it using gradient descent.
The challenge lies in the exponentially many candidates of $\mathcal{T}(D)$, which is computationally infeasible to naively enumerate.
Fortunately, we will see that the Matrix-Tree Theorem~\cite{tutte1984graph,koo2007structured} permits efficient calculation of the summation we need.

\subsubsection{Matrix-Tree Theorem}
Before we dive into the details of the Matrix-Tree Theorem, we have to give enough credits to~\citet{tutte1984graph, koo2007structured} as we are applying their proposed theorem.
The probability of the reference tree $\bar T$ , which is to be optimized, can be defined as:
\begin{equation}
    \mathbb{P}(\bar T) = \frac{s(\bar T)}{Z(\theta)}, \quad Z(\theta) = \sum_{T\in \mathcal{T}(D)} s(T)
    \label{eq:p}
\end{equation}
$Z(\theta)$ is also known as the partition function.
The numerator $s(T)$ of any tree $T$ is defined to be:
\begin{equation}
    s(T) = \prod_{(h,m,r)\in T} \exp (\theta_{h,m,r}),
\end{equation}
With this definition, the score is merely the product of corresponding cells in $\exp(\theta)$ ($\theta$ from eq.~(\ref{eq:encoder})).

Next, we need to find an efficient way to compute the partition function as there are exponentially many candidate trees. The first step is to calculate the exponential matrix $A_{h,m,r}$ from eq.~(\ref{eq:encoder}):
\begin{equation}
  A_{h,m,r}(\theta) =
    \begin{cases}
      0, & \text{if $h\geq m$} \\
      \exp(\theta_{h,m,r}), & \text{otherwise}
    \end{cases}
    \label{eq:pre_potential}
\end{equation}
Note that the first $0=\exp(-inf)$ condition applies to all $h\geq m$ cells as $\theta$ is an upper-triangular matrix described earlier.
To account for edge labels, we have to marginalize $r$ out :
\begin{equation}
  A_{h,m}(\theta) = \sum_r A_{h,m,r}(\theta)
    \label{eq:potential}
\end{equation}
Now, we are ready to calculate $Z(\theta)$.
The first step is to calculate the graph Laplacian matrix, which is the difference between the degree matrix and adjacency matrix:
\begin{equation}
  L_{h,m}(\theta) =
    \begin{cases}
      \sum_{i'=1}^n A_{i',m}(\theta), & \text{if $h=m$} \\
      -A_{h,m}(\theta), & \text{otherwise}
    \end{cases}       
\end{equation}
Then the minor\footnote{A minor $L^{(x,y)}$ is the determinant of a submatrix constructed by removing the $x$-th row and $y$-th column of $L$.} $L^{(0,0)}(\theta)$ is equal to the sum of the weights of all directed spanning trees rooted at the dummy root utterance~\cite{tutte1984graph}:
\begin{equation}
    Z(\theta) = \det(\hat L),
    \label{eq:det}
\end{equation}
whre $\hat L$ is defined to be the submatrix constructed by removing the first row and column from $L$.
The computational complexity of eq.~(\ref{eq:det}) is determined by the determinant operation, which is $O(n^3)$.
While the cubic time complexity might seem scary at the first glance, it does not incur significant computational overhead in our experiments, where the time to compute the determinant is negligible (< 1\%) compared to BERT encoding in eq.~(\ref{eq:independent_score}).

\begin{figure*}[]
\hspace*{-10mm} 
\includegraphics[width=1.1\textwidth]{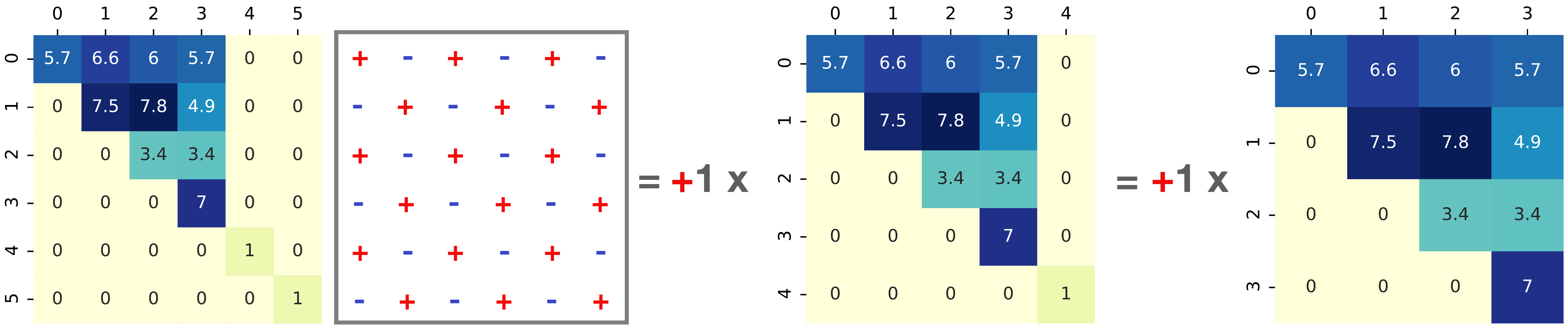}
\caption{This is an efficient padding for calculating batch determinant. The original $4\times 4$ matrix is expanded to a $6\times6$ (leftmost) one for padding. Note that the last two diagonal elements are all ones. The second matrix encodes the coefficients for multiplying sub-matrices. After a series of cofactor expansions, we can see that the determinant of the padded $6\times 6$ matrix is equivalent to the original unpadded $4\times4$ matrix.}
\label{fig:batch_det}
\vspace{-3mm}
\end{figure*}

\subsubsection{Efficient GPU Implementation}
The equations derived so far work well for a single training instance. However, it becomes problematic if we want to perform batchwise training on GPUs, which was not addressed in~\citet{koo2007structured}.
The main challenge is the variable-length padding. In particular, we have to calculate batchwise determinants in eq.~(\ref{eq:det}) with different sizes of $\hat L$. The naive option is to pad the extra rows and columns with zeros. Unfortunately, this would result in a singular matrix and give erroneous partition results. To circumvent the padding issue, we can use the cofactor expansion formula. Concretely, all the diagonal elements of the padding part should be 1, while others should be 0. We illustrate the padding strategy in Figure~\ref{fig:batch_det}. Note that this strategy holds whether the size of $\hat L$ is odd or even.

\subsection{Optimization of Tree}
Since we are given the reference tree $\bar T$, we can directly maximize the log probability of eq.~(\ref{eq:p}) using any gradient-descent based algorithms, which is also equivalent to minimizing the KL-divergence between the predicted and reference tree distributions.

\subsection{Inference of Tree}
\label{sec:inference}
There is a well-known algorithm - Chiu-Liu-Edmonds (CLE)~\cite{edmonds1967optimum,chu1965shortest} that can find the directed spanning tree $\tilde T$ with maximum weight given $A(\theta)$ derived in eq.~(\ref{eq:pre_potential}). However, we cannot directly apply the CLE algorithm as the original version does not accept labeled trees. To solve this problem, we have to first pick the highest-scoring relation for each edge $A'_{h,m}=\max_r A_{h,m,r}$ to get $A'\in \mathbb{R}^{(n+1)\times (n+1)}$. Now we can feed this standard form into the CLE algorithm: $\tilde T = \mathrm{CLE}(A')$. The correctness of this approach can be proved easily by contradiction: suppose the optimal tree includes one edge that is not the highest-scoring one among $\{A_{h,m,r}\}_{r=1}^{17}$, we can always substitute that edge with the highest-scoring one to get a better tree (contradiction). Note that for a pair of utterances, we only allow one direction of link ($\theta$ is strictly upper triangular) so the CLE algorithm in fact degenerates to its undirected version known as Prim/Kruskal's algorithm~\cite{prim1957shortest,kruskal1956shortest}.

\section{Experiments}
\label{sec:exp}
\subsection{Datasets}
There are two datasets for us to train the discourse parser, one of which is the STAC~\cite{asher2016discourse} corpus, which is a multi-party dialogue corpus collected from an online game, and the other is the Ubuntu IRC corpus~\cite{li2020molweni}, which compiles technical discussions about Linux. The differences between these two datasets were analyzed in~\citet{liu-chen-2021-improving}, where the takeaway messages are: 1) there is no significant difference in their average EDU numbers, 2) the lexical distributions are significantly different sharing only a small portion of common
tokens, 3) relation distributions are similar.

\subsection{Hyperparameters}
Following~\citet{liu-chen-2021-improving}, we use the Roberta-Base uncased pretrained checkpoint for a fair comparison. The max utterance length is set to 28. The initial learning rate is set to 2e-5 with a linear decay to 0 for 4 epochs. The batch size is 4. The first 10\% of training steps is the warmup stage. For all baselines using large pretrained models, we always use the same model checkpoint and tune the learning rate and batch size for them for a fair comparison.

\subsection{Metrics}
We follow the baselines to use two metrics for evaluation:
\begin{compactitem}
    \item Unlabeled Attachment Score (UAS): We only care about the existence of a discourse link. In other words, discourse relations do not affect the results. (Also known as Link F1 score)
    \item Labeled Attachment Score (LAS): It is much harder as it requires both discourse links and relations to be correct. We focus primarily on this metric since it is more informative and is used in downstream applications. (Also known as Link \& Rel F1 score)
\end{compactitem}

\begin{table*}\centering
\hspace*{-1em}\begin{tabular}{@{}lccccccccccc@{}}\toprule
& \multicolumn{2}{c}{\text{STAC / STAC}} & \phantom{}& \multicolumn{2}{c}{\text{MOL / MOL}} &
\phantom{} & \multicolumn{2}{c}{\text{STAC / MOL}} & \phantom{} & \multicolumn{2}{c}{\text{MOL / STAC}} \\
\cmidrule{2-3} \cmidrule{5-6} \cmidrule{8-9} \cmidrule{11-12}
& \text{UAS} & \text{LAS} && \text{UAS} & \text{LAS} && \text{UAS} & \text{LAS} && \text{UAS} & \text{LAS} \\ \midrule
MST~\cite{afantenos2015discourse} & 69.6 & 52.1 && 69.0 & 48.7 && 61.5 & 24.0 && \textbf{60.5} & 14.8 \\
ILP~\cite{perret2016integer} & 69.0 & 53.1 && 67.3 & 48.3 && 57.0 & 24.1 && 60.4 & 14.5 \\
Deep-Seq.~\cite{shi2019deep} & 73.2 & 54.4 && 76.1 & 53.3 && 53.5 & 21.6 && 42.7 & 15.7 \\
Hierarchical~\cite{liu-chen-2021-improving} & 73.1 & 57.1 && 80.1 & 56.1 && 60.1 & 32.1 && 48.9 & 26.8 \\
Struct-Aware~\cite{wangstructure} & 73.4 & 57.3 && 81.6 & 58.4 && 57.0 & 32.9 && 44.7 & 26.1 \\
This Work & \textbf{74.4} & \textbf{59.6} && \textbf{83.5} & \textbf{59.9} && \textbf{64.5} & \textbf{38.0} && 50.6 & \textbf{31.6} \\
\bottomrule
\end{tabular}
\caption{STAC / MOL means the training dataset is STAC and the testing dataset is MOL. LAS is the harder setting used for downstream applications. Results are the average of three runs. Note that speaker information is still used in this set of experiments, except our parser does not need to model their relations explicitly as described in \S~\ref{sec:feature}.}
\label{tab:result}
\end{table*}

\subsection{Main Results}
We present the results in Table~\ref{tab:result}. The left part of the table focuses on in-domain training and testing, which is the standard setting. Bearing in mind that discourse parsers are often used as the first stage of downstream applications, we follow~\cite{liu-chen-2021-improving} to benchmark the performance of all parsers in the cross-domain setting. Note that this is an extremely challenging setting as the domains are completely different (gaming vs Linux technical forum).
\paragraph{In-domain}
We first take a look at the in-domain results. Our proposed parser is the best among all parsers, surpassing the previous state-of-the-art by 2.3 on STAC and 1.5 (F1 scores) on Molweni under the LAS setting. The trend is similar for the UAS setting. We also want to highlight the improved performance can NOT be attributed to using a pretrained language model as~\emph{Struct-Aware} and~\emph{Hierarchical}~\cite{liu-chen-2021-improving} both utilize the same or comparable pretrained model.
\paragraph{Cross-domain}
We shift gear to the cross-domain setting where the parser is trained on one dataset and tested on the other~\cite{liu-chen-2021-improving}. We can see that our parser is the best under the LAS setting, substantially outperforming the best candidates by 5.1 on STAC/MOL and 4.8 points on MOL/STAC.
However, the best-performing model under the UAS setting is the oldest model~\cite{afantenos2015discourse,perret2016integer}. This can be explained by the inclination of a large pretrained model to overfit on the training domain, which was corroborated by~\citet{liu-chen-2021-improving} as well. Readers might wonder why the same phenomenon does not happen under the UAS setting of STAC/MOL, and the speculated reason is STAC has a much larger linguistic diversity~\cite{liu-chen-2021-improving}, thereby alleviating the model overfitting issue. In other words, we might want to train the dialogue discourse parser on a linguistically diverse dataset if the goal is domain generalization.

\begin{figure}[]
\includegraphics[width=\linewidth]{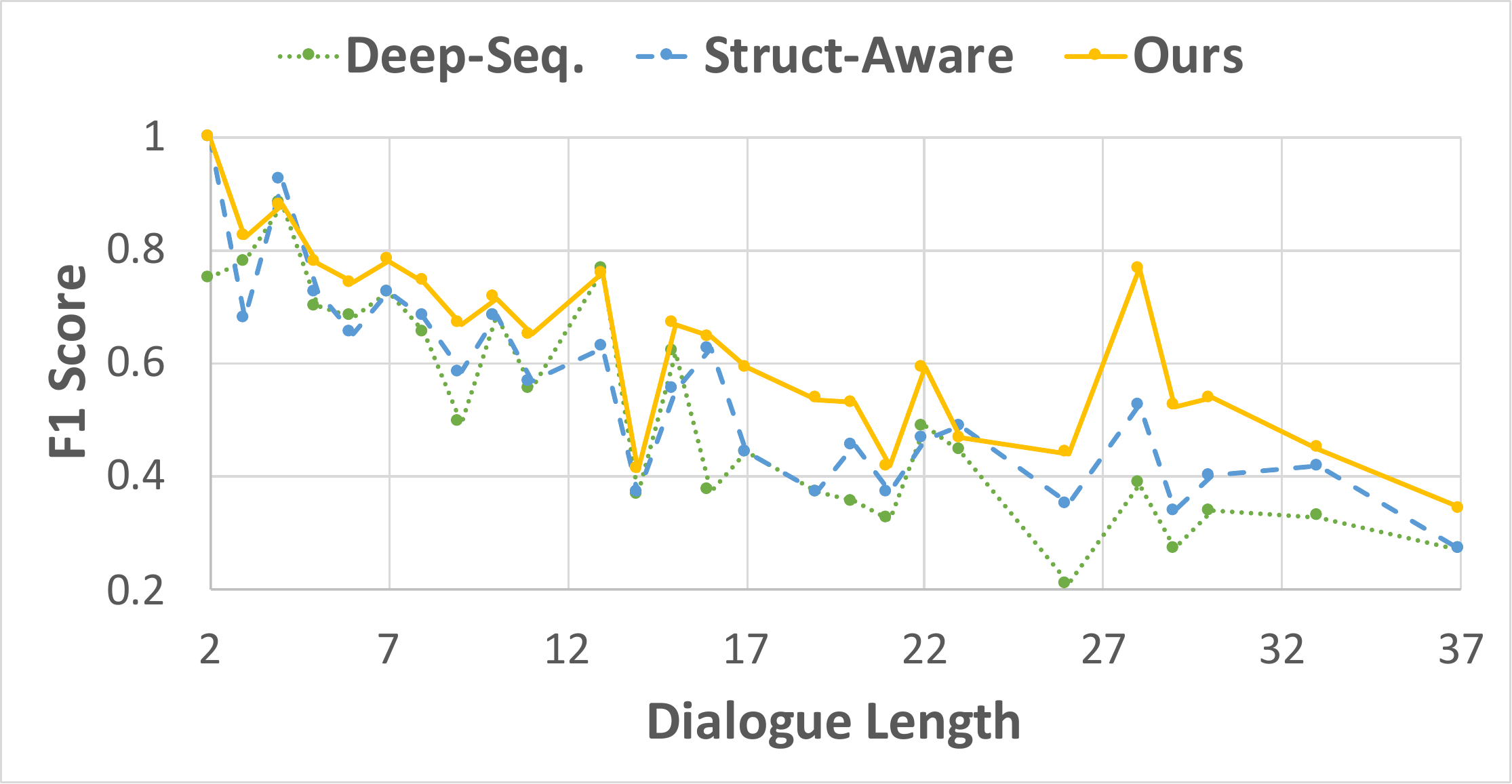}
\caption{Parsing performance w.r.t dialogue lengths. As we can see, the performance difference is larger when the dialogue becomes longer, demonstrating the length robustness of our parser.}
\label{fig:length}
\end{figure}

\subsection{Additional Analyses}
\paragraph{Dialogue Length Robustness}
We hypothesize that our parser is likely to perform better when the dialogue becomes longer, and the reason is that our parser models the overall dialogue structure using tree distributions. This lowers the burden of the parser to predict long-range links.
We focus on the in-domain setting and plot the results in Figure~\ref{fig:length}. As we can see, the performance of our parser drops less than baselines when the dialogue becomes longer, highlighting the benefit of global structured learning and inference.

\begin{figure}[]
\includegraphics[width=\linewidth]{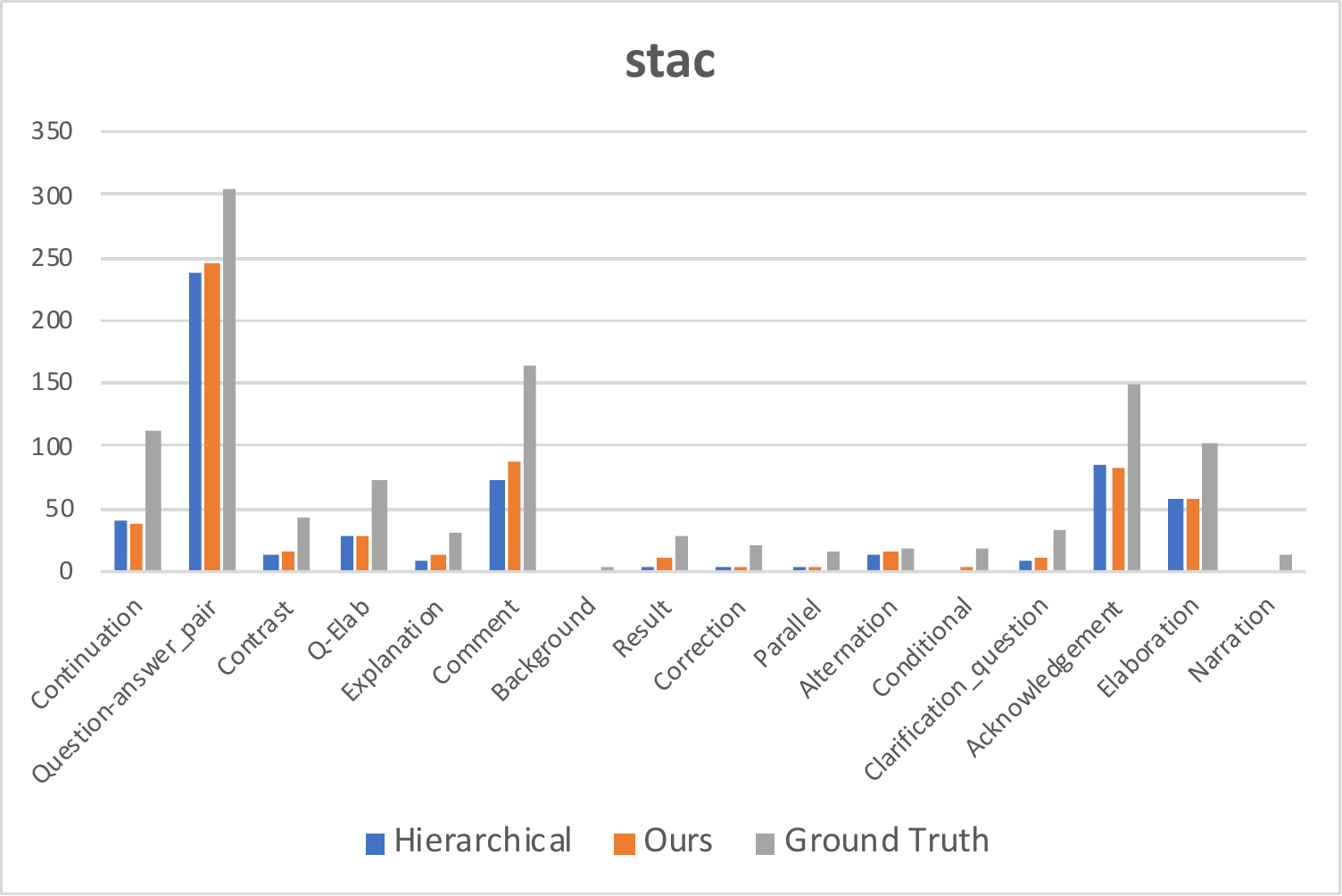}
\caption{STAC relation performance breakdown.}
\label{fig:stac_breakdown}
\end{figure}

\begin{figure}[]
\includegraphics[width=\linewidth]{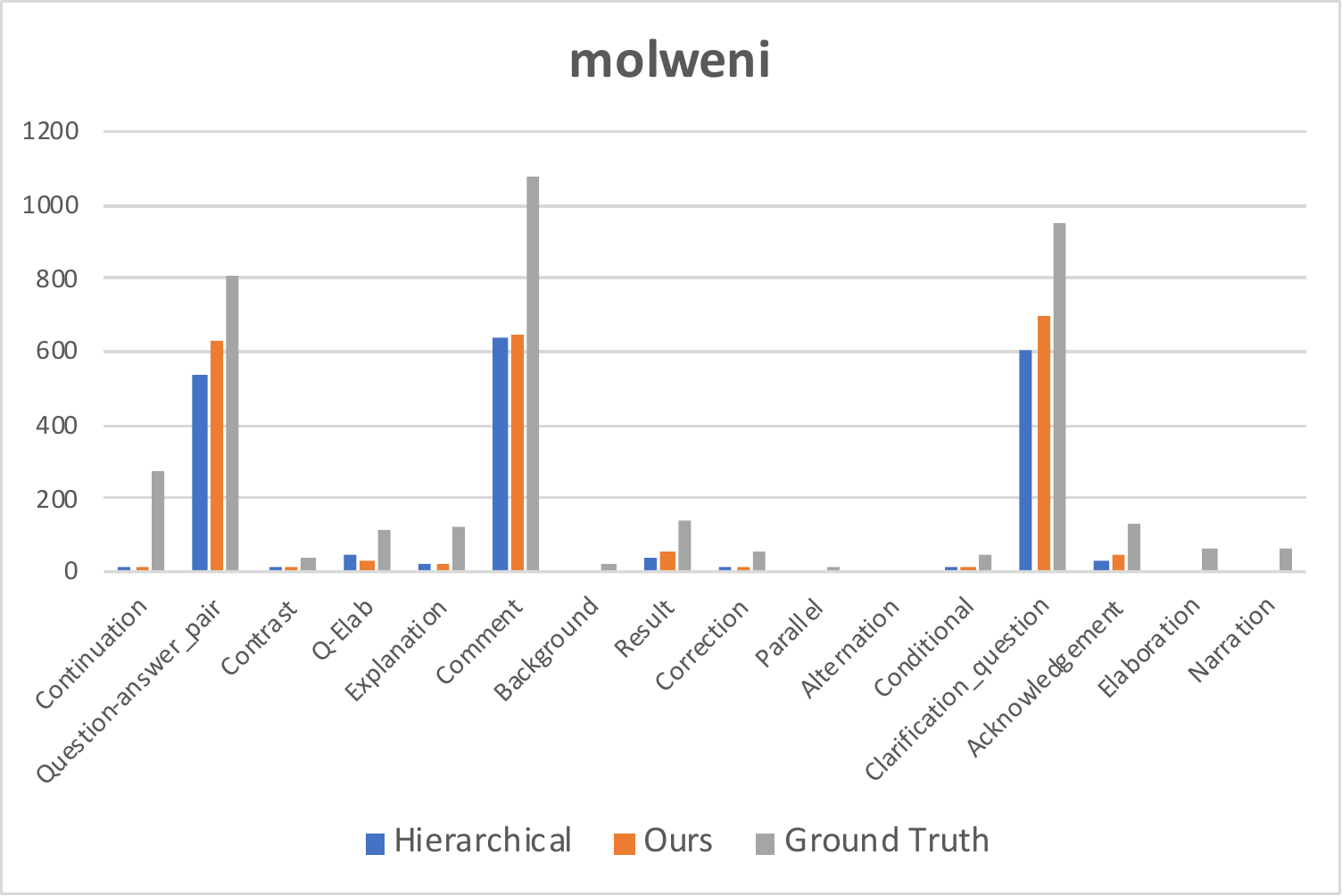}
\caption{Molweni relation performance breakdown.}
\label{fig:molweni_breakdown}
\end{figure}

\paragraph{Relation Performance Breakdown}
In order to know what kinds of relations benefit the most from our proposed parser, we count the number of correct relation predictions and plot them in Figure~\ref{fig:stac_breakdown} and~\ref{fig:molweni_breakdown}.\footnote{Note that this implies the link predictions of these correct relations are also correct.} The baseline parser we compare with is the Hierarchical model~\cite{liu-chen-2021-improving} as it can be viewed as the non-structured version of our parser with the same pretrained model backbone. We can see that our parser outperforms the baseline on certain relations like~\emph{Comment} and~\emph{QA pair} on STAC and ~\emph{QA pair} and~\emph{Clarification Question} on Molweni.
However, there is still a large room for improvement as demonstrated by the gap between our parser and the ground truth. Another observation is that both parsers struggle to predict low-resource relations, marking an important direction for future work.

\paragraph{Speaker and Turn Feature Robustness}
We experiment with removing speaker and turn information used in baselines. The performance drop (LAS of STAC) of our parser (59.6 $\to$ 54.4) is less than that of the best baseline (Struct-Aware) (57.3 $\to$ 47.8), demonstrating the robustness of our parser.

\section{Related Work}
\paragraph{Discourse Parsing}
As discussed in the Introduction section, there are three types of discourse parsing formalisms: RST~\cite{mann1988rhetorical}, PDTB~\cite{prasad2008penn}, and SDRT~\cite{lascarides2008segmented, asher2016discourse}.
For the first two tasks, there are transition-based~\cite{li2014text,braud2017cross,yu2018transition} and CKY-based methods~\cite{joty2015codra,li2016discourse,liu2017learning} in the literature.
In this work, we assume that the EDUs are already given. In practice, there are papers working on segmenting EDUs~\cite{subba2007automatic,li2018segbot} before feeding them to the discourse parser.

\paragraph{Dialogue Disentanglement}
Clustering utterances in a conversation into threads is studied extensively by previous work~\cite{shen2006thread,elsner2008you,wang2009context,elsner2011disentangling,jiang2018learning,kummerfeld2018large,zhu2020did,li2020dialbert,yu2020online}. 
They predict the~\emph{reply-to} links independently and run a connected component algorithm to construct the threads. This is similar to the UAS setting in this work.

\paragraph{Structured Learning Algorithms}
Natural language is highly structured suggesting that the introduction of structural bias will facilitate learning. Previous work have studied dependency-tree like structures extensively~\cite{koo2007structured,mcdonald2005non,mcdonald2007complexity,niculae2018sparsemap,paulus2020gradient}. Several works propose to incorporate such inductive bias into intermediate layers of modern NLP models~\cite{kim2017structured,chen2017improved,liu2018learning,choi2018learning}. In our work, the induced structure is not only implicitly learned, it is also used to directly decode the labeled tree structure, which is our ultimate goal.

\paragraph{Dependency Parsing}
Our work can also be viewed as extending token-level dependency parsing~\cite{mel1988dependency,koo2007structured,smith2008dependency,koo2010efficient,chen2014fast,dozat2016deep,qi2019universal,choi2011getting} to utterance-level. Another important difference is that our tree is labeled, which means we have to additionally predict the type of tree edges.

\section{Conclusion}
In this paper, we propose a principled method for dialogue discourse parsing. From the encoding side, we introduce a structurally-encoded adjacency matrix followed by the matrix-tree theorem, which is used to holistically model all utterances as a tree. From the decoding side, we apply the modified CLE algorithm for maximum spanning tree induction. Our method achieves state-of-the-art performance on two benchmark datasets. We also benchmark the cross-domain parser performance, and find our parser performs the best in the most-commonly used and harder LAS setting. We believe that the techniques described in this work pave the way for more structured analyses of dialogue and interesting research problems in the field of dialogue discourse parsing.

\section*{Acknowledgment}
The authors acknowledge the support from Boeing (2019-STU-PA-259).
\bibliography{acl}
\bibliographystyle{acl_natbib}

\end{document}